%% file: main.tex
\documentclass[conference]{IEEEtran}

\usepackage{graphicx}
\usepackage{amsfonts}
\usepackage{amsmath}
\usepackage{xspace}
\usepackage{xcolor}
\usepackage{booktabs}
\usepackage{multirow}
\usepackage{enumitem}
\usepackage{algorithm2e}
\usepackage[noabbrev, capitalize]{cleveref}
\usepackage{pgfplots}
\pgfplotsset{compat=1.17}
\usepackage{tikz}
\usetikzlibrary{positioning, arrows.meta, calc, shapes.geometric, arrows}
\usepackage{authblk}
\newcommand{\sys}{\textit{SuperPure}\xspace}
\newcommand{\sysp}{\textit{SuperPure+}\xspace}
\newcommand{\ours}{SuperPure\xspace}

\title{SuperPure: Efficient Purification of Localized and Distributed Adversarial Patches via Super-Resolution GAN Models}

\author[1]{Hossein Khalili}
\author[1]{Seongbin Park}
\author[1]{Venkat Bollapragada}
\author[1]{Nader Sehatbakhsh}
\affil[1]{University of California, Los Angeles (UCLA)}

\date{}  

\begin{document}
\maketitle

\begin{center}
\textbf{Preprint. This work has been submitted to arXiv.}
\end{center}

\begin{abstract}
As vision-based machine learning models are increasingly integrated into autonomous and cyber-physical systems, concerns about (physical) adversarial patch attacks are growing. While state-of-the-art defenses can achieve certified robustness with minimal impact on utility against highly-concentrated localized patch attacks, they fall short in two important areas: \textbf{\textit{(i)}} they are vulnerable to low-noise \textit{distributed} patches where perturbations are subtly dispersed to evade detection or masking, as shown recently by the DorPatch~\cite{he2024dorpatch} attack; \textbf{\textit{(ii)}} they are extremely time and resource-consuming, making them impractical for latency-sensitive applications.

To address these challenges, we propose \textit{SuperPure}, a new defense strategy that combines pixel-wise adversarial masking and GAN-based super-resolution. It is robust against both distributed and localized patches while achieving low inference latency. Extensive evaluations using ImageNet and two standard classifiers (ResNet and EfficientNet) show that SuperPure: \textbf{(i)} improves robustness against localized patches by >20\% and clean accuracy by ~10\%; \textbf{(ii)} achieves 58\% robustness against distributed patch attacks (vs. 0\% for PatchCleanser); \textbf{(iii)} reduces defense latency by 98\%. SuperPure is robust to patch sizes and white-box attacks. Code is open-source.
\end{abstract}

\section{Introduction}
\input{Sections/Intro}

\section{Background}
\input{Sections/Related}

\section{Problem Formulation}
\input{Sections/Threat-Model}

\section{Method}
\label{sec:sys}
\input{Sections/Method}

\section{Results}
\label{sec:results}

\input{Sections/new_results_5}

\section{Ablation Studies}
\label{sec:ablations}
\input{Sections/Ablations}

\section{Conclusions}
In this paper, we proposed a new model-agnostic defense method against both singular and distributed patch attacks. \sys utilizes discrepancies between the outputs of a reconstructed image using a super-resolution GAN and the original input to mask adversarial regions; for smaller, less perceptible patches, \sys includes an enhancement step (\sysp) to filter adversarial perturbations missed by the masking process. Through extensive experiments, we demonstrated the superior robustness of our method compared to prior work, showcasing its ability to defend effectively against a variety of patch-based adversarial attacks.  

\bibliographystyle{ieeetr} 
\bibliography{references}

\appendix
\input{Sections/appendix}

\end{document}

%% file: Sections/Intro.tex





         

Deep learning models have achieved remarkable success in various computer vision tasks including image classification, object detection, and semantic segmentation \cite{krizhevsky2012imagenet, he2016deep, dosovitskiy2021an}. However, they are highly vulnerable to adversarial attacks, which involve adding perturbations to input data in order to mislead models into making incorrect predictions \cite{szegedy2014intriguing, goodfellow2015explaining}. Among these, adversarial patches—large perturbations confined to a localized region—pose a significant and practical threat due to their effectiveness and ease of deployment \cite{brown2017adversarial, karmon2018lavan}. Specifically, they pose a serious threat to real-world vision applications such as autonomous driving and security systems, where attackers can physically print out and place a patch in the environment to manipulate the model’s output \cite{eykholt2018robust, liu2018dpatch}.

    


Various defense strategies have been proposed to counter adversarial patches~\cite{jing2024pad,chiang2020certified,xiang2021patchguard,xiang2022patchcleanser,xiang2024patchcure,mao2022defending}. A common approach involves eliminating adversarial samples through various forms of masking and image transformations. Currently, the most popular and effective defense method is one developed by Xiang \textit{et al.} called PatchCleanser~\cite{xiang2022patchcleanser}. This certified defense technique identifies and masks suspicious areas in an image, using randomized smoothing methods to provide theoretical guarantees. However, while PatchCleanser performs well against localized patches of a specific size, its effectiveness diminishes when it encounters advanced \textit{distributed} attacks that spread perturbations to escape detection. Exploiting this vulnerability, Tang \textit{et al.} introduced DorPatch~\cite{he2024dorpatch}, a dispersed and occlusion-robust adversarial patch attack that distributes perturbations across multiple regions. Importantly, the misclassified results from adversarially patched samples created by DorPatch can obtain certification from PatchCleanser, leading to a false sense of security in the guaranteed predictions. These findings highlight the urgent need for the development of effective defenses to counter such attacks.

\begin{figure*}
         \centering
         \includegraphics[width=1.70\columnwidth]{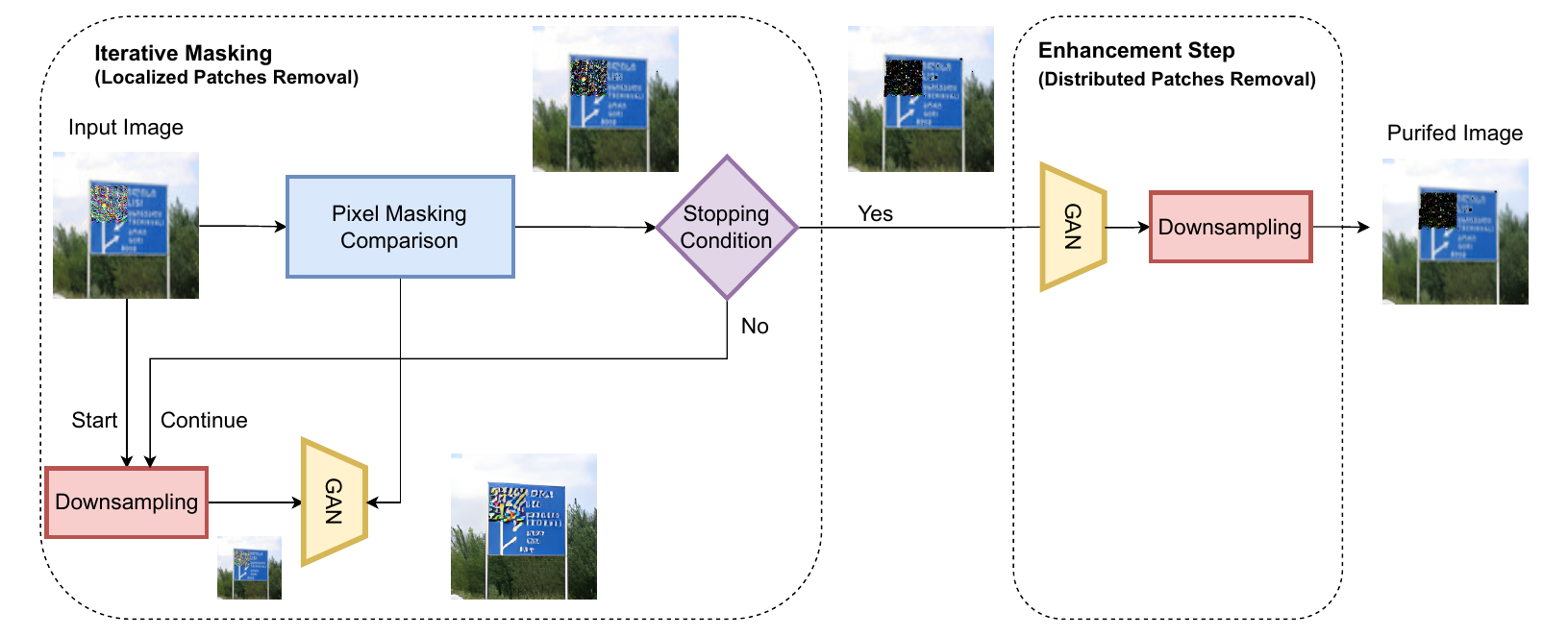}
         \caption{\ours pipeline: at each iteration, we downsample, GAN-upsample, and mask high-error pixels. If newly masked pixels exceed the threshold, we repeat. At the end, a final upsample--downsample “enhancement” removes subtle perturbations.} 
         \label{fig:overview}
\end{figure*}

Apart from concerns about \textit{robustness} against distributed attacks, the \textit{computation overhead} of current defense mechanisms has also been a growing concern. This is particularly important when considering latency-sensitive applications such as image classification for autonomous systems, where the solution has to be both robust and lightweight. Despite recent efforts to further improve the computational efficiency~\cite{xiang2024patchcure}, the overhead remains quite high—on the order of \textit{tens} of seconds per image for state-of-the-art methods~\cite{xiang2024patchcure, xiang2022patchcleanser}, as we will demonstrate in this paper. 

This paper introduces \textit{\sys{}}, a novel iterative defense mechanism designed specifically to counter adversarial patch attacks, addressing critical robustness and latency constraints. At its core, \textit{\sys{}} iteratively masks adversarial patches by combining low-pass downsampling, GAN-based super-resolution upsampling, and precise pixel-level comparisons to identify and eliminate both localized and distributed adversarial signals.

At each iteration, \textit{\sys{}} first downsamples the input image. This step acts as a low-pass filter, attenuating high-frequency adversarial perturbations while retaining essential image content~\cite{xu2018feature, shannon1949communication, tancik2020fourier}. The primary structure and semantics of the image remain largely unaffected due to inherent redundancy and spatial correlation typically observed in natural images~\cite{torralba2003statistics}. The iterative non-linear GAN-based upsampling further disrupts adversarial features, progressively restoring clean image regions.

An important insight was that simple upsampling with a GAN would \underline{not} be sufficient. Instead, we propose a novel method called \textbf{\textit{GAN-guided pixel masking.}} Specifically, we first upsample the image using the GAN to introduce \textit{non-linearity} to the processing pipeline, and then \textit{compare} this newly generated image with the original image. The key insight is that, since the super-resolution model (GAN) is trained on natural (clean) images, \textit{it affects the pixels related to the patches more than the clean pixels when creating a new image.} We then implement an additional non-linear step called pixel masking, in which the pixels of the newly generated image are compared with those of the original image on a pairwise basis. Any pixels that differ significantly from the original, based on a predefined threshold, are masked. This approach allows our method to eliminate pixels that are more influenced by the down-sampling and super-resolution processes. The important takeaway is that pixels associated with the patch are much more likely to be among those that differ. 

The second purpose of GAN-guided pixel masking is to \textit{improve clean accuracy}. Simple downsampling and upsampling alone can lead to significant information loss, which in turn reduces clean accuracy; hence, a more sophisticated scheme is used for upsampling.


This \textit{down-up sampling plus masking} process continues until the percentage of newly masked pixels drops below a small threshold. We find that this method is more effective when changes are gradual. Additionally, to guard against less noticeable, distributed patches, we include an \textit{enhancement step} that involves upsampling the image using super-resolution, followed by downsampling it back to its original size.  Figure~\ref{fig:overview} illustrates our scheme.

To minimize computation overhead, \sys utilizes several techniques. Primarily, it employs a lightweight GAN-based super-resolution model for the upsampling step. It also utilizes an adaptive stopping condition. In Section \ref{sec:sys}, we outline the details.
Through extensive experiments on the ImageNet dataset \cite{deng2009imagenet} using state-of-the-art classifiers, Resnet and EfficientNet, we demonstrate that our method significantly enhances robustness against adversarial patch attacks of both singular and distributed varieties. We compare our method to the widely popular method, PatchCleanser~\cite{xiang2022patchcleanser}, and the most recent defense mechanism in this domain, PAD~\cite{jing2024pad}. Compared to state-of-the-art, \sys improves the robustness against conventional localized patches by more than 20\%, on average, while also improving top-1 clean accuracy by almost 10\%. \sys also achieves 58\% robustness against distributed patch attacks while PatchCleanser~\cite{xiang2022patchcleanser} is completely vulnerable to this attack (0\% robustness). Lastly, \sys also decreases the end-to-end latency by over 98\%.


In summary, our work presents the following contributions:
\begin{enumerate}
    \item We present \sys, a novel defense mechanism against white-box patch attacks of both singular and distributed varieties that is compatible with various image classifiers.
    \item We introduce multiple techniques to reduce the computation complexity of our approach without significantly impacting the accuracy and/or robustness of the defense. 
    \item We evaluate \sys on standard datasets and classifiers and compare our method against state-of-the-art methods.
    \item We provide ablation and sensitivity studies to provide more insights about our method and highlight its effectiveness. 
    \item We open-source our source code. 
\end{enumerate}





%% file: Sections/Related.tex
\subsection{Adversarial Patch Attacks} 
Given a model $M$ that produces an output $M(x)$ given a sample $x$, adversarial attacks involve finding a modified sample $\Tilde{x}$ such that $M(\Tilde{x}) \neq M(x)$ 
\cite{szegedy2014intriguing, carlini2017towards, barreno2010security, biggio2013evasion, papernot2016limitations}. A common type of attack involves adding imperceptible perturbations throughout the whole image, designed to subtly distort the image without drawing human attention \cite{croce2020minimally, goodfellow2015explaining, madry2018towards}. 

In contrast, \textbf{adversarial patch attacks} add conspicuous perturbations to a restricted area of the image called a patch \cite{hayes2018visible, karmon2018lavan}. These patches are especially relevant in physical settings where they can be applied as stickers or printed patches \cite{ brown2017adversarial, wei2022adversarial, eykholt2018robust}. Early papers assumed a white-box threat model, where adversaries have access to the internal details of the target classifier. In contrast, PatchAttack introduced a reinforcement learning-based method for generating adversarial patches in a black-box setting \cite{yang2020patchattack}. 

Unlike singular patches that are localized to one specific region, RP2 \cite{eykholt2018robust} uses a distributed adversarial patch, which is harder to locate and cover. More recently, researchers proposed Distributed and Occlusion-Robust Adversarial Patch (DorPatch), which uses group lasso on patch masks, image dropout, density regularization, and structural loss to produce a distributed and occlusion-robust patch for real-world deployment \cite{he2024dorpatch}. 

There have been adversarial patch attacks proposed in other domains such as object detection \cite{liu2018dpatch, lee2019physical}, but in this paper, we focus on image classifiers as our target model. 

\subsection{Defense Mechanisms}
Initial defenses against adversarial patches include Digital Watermarking \cite{hayes2018visible}, which involves masking unnaturally dense regions in the saliency map, and Local Gradient Smoothing \cite{naseer2019local}, which regularizes gradients in the estimated noisy regions. However, Chiang \textit{et al.} showed that these defenses can be bypassed using adaptive white-box attacks, and proposes the first certifiable defense against patch attacks by extending interval bound propagation (IBP) defenses \cite{chiang2020certified}. Clipped BagNet (CBN) \cite{zhang2020clipped} uses clipped BagNet, a classification model with small receptive fields, for certified robustness. Levine and Feizi extends randomized smoothing robustness schemes by leveraging the constrained nature of patch attacks to derive larger, deterministic robustness certificates. However, these works are overshadowed by PatchGuard \cite{xiang2021patchguard} in terms of performance.

PatchGuard utilizes small receptive fields in CNNs to prevent adversarial patches from influencing classification. However, this approach is less effective for larger models with wider receptive fields and requires significant architectural modifications, limiting its broader applicability. PatchCleanser \cite{xiang2022patchcleanser} employs two rounds of pixel masking on the input image to neutralize the effect of adversarial patches. PatchCure is based on the same principles as PatchCleanser but is optimized for efficiency \cite{xiang2024patchcure}. However, unlike PatchCleanser, PatchCure requires retraining the underlying classifiers to integrate its defense mechanisms, introducing additional computational overhead and limiting its practical deployment flexibility. Both PatchCleanser and PatchCure are vulnerable to disturbed patch attacks \cite{he2024dorpatch}.

Many patch localization-based defenses have also been proposed. SentiNet utilizes techniques from model interpretability and object detection to detect potential attacks \cite{chou2020sentinet}. Jedi utilizes input entropy analysis to detect and remove adversarial patches \cite{tarchoun2023jedi}. PatchZero \cite{xu2023patchzero} detects adversarial pixels using a patch detector and repaints them with the mean color of the surrounding pixels. Similarly, PAD \cite{jing2024pad} leverages mutual information and recompression to locate and remove adversarial patches, effectively addressing challenges related to patch appearance, shape, size, location, and quantity without needing prior attack knowledge.

There have also been research done utilizing adversarial training \cite{goodfellow2015explaining}, which is widely used to mitigate attacks based on adversarial samples, to mitigate patch attacks. These works \cite{rao2020adversarial, wu2020defending, addepalli2021efficient} propose novel training schemes to boost robustness, but they require retraining and/or incur high computational overhead. In this work, we focus on preprocessing defenses that do not require retraining of the target model.

\subsection{Image Super Resolution}
Image Super-Resolution (SR) is a long-standing problem in computer vision, aiming to recover a high-resolution (HR) image from its low-resolution (LR) counterpart. Traditional interpolation methods, such as bicubic and bilinear interpolation, often fail to reconstruct fine details and high-frequency textures, resulting in blurred images.

With the rise of deep learning, Convolutional Neural Networks (CNNs) have become predominant in SR tasks. Dong et al. \cite{dong2014learning} introduced the Super-Resolution Convolutional Neural Network (SRCNN), pioneering the use of deep learning for SR. Following this, Kim et al. \cite{kim2016accurate} proposed the Very Deep Super-Resolution (VDSR) network, utilizing a much deeper architecture and residual learning to enhance performance.

In addition to this, Generative adversarial networks (GANs) [14] have become a popular approach, often used as a loss function to push the results closer to the natural image manifold, improving perceptual quality. Ledig et al. \cite{ledig2017photo} proposed the Super-Resolution Generative Adversarial Network (SRGAN), which introduced an adversarial loss and a perceptual loss based on high-level feature mappings from a pre-trained VGG network. This approach enabled the generation of more photo-realistic images with sharper details. However initial methods often struggle to perform well on real-world images. This was due to the fact that standard blurs or filters often did not capture the complex intricacies of image degradation. Building upon SRGAN, Wang et al. \cite{wang2018esrgan} developed the Enhanced SRGAN (ESRGAN), which introduced the Residual-in-Residual Dense Block (RRDB) to facilitate training of deeper networks without degradation. ESRGAN also replaced the standard discriminator with a relativistic average discriminator to provide stronger supervision and better gradient flow, resulting in superior perceptual quality.

\subsection{Super-Resolution Defenses}

Recent works have utilized super-resolution (SR) to eliminate adversarial noise by projecting low-quality inputs back onto a manifold of natural images. Mustafa~\textit{et al.}~\cite{mustafa2019image} introduced an SR-based approach to remove subtle \(\ell_p\)-bounded perturbations, restoring clean performance at modest overhead. Meanwhile, DiffPure~\cite{nie2022diffpure} applies diffusion-based sampling to expunge mild adversarial artifacts, also yielding high-quality outputs under small or dispersed corruptions. However, \textbf{both} SR- and diffusion-based defenses commonly assume lower-amplitude distortions; large, localized patch signals often remain partially intact—sometimes even sharpened—after SR. Consequently, these methods fail to robustly handle overt patch attacks without dedicated detection and masking strategies.

%% file: Sections/Threat-Model.tex
In this section, we clearly define the objectives of the attacker and defender considered in this paper, focusing explicitly on two primary adversarial scenarios: \textit{localized} and \textit{distributed} patches.

\subsection{Attack Objective}\label{sec:att-obj}

\subsubsection{Attacker Goal}\label{sec:att-goal}
The adversary seeks to mislead a classifier by embedding adversarial patches into input images. Formally, consider an original clean image $\mathbf{x} \in \mathbb{R}^{H \times W \times C}$, where $H$, $W$, and $C$ represent its height, width, and number of channels, respectively. An adversarial patch introduces an additive perturbation $\boldsymbol{\delta} \in \mathbb{R}^{H \times W \times C}$, resulting in an adversarial image:
\begin{equation}
    \mathbf{x}_{\text{adv}} = \mathbf{x} + \boldsymbol{\delta}.
\end{equation}

We specifically consider two distinct types of adversarial patches:

\begin{itemize}
    \item \textbf{Localized patches:} Perturbations are concentrated within a single, contiguous region $\mathcal{P}_L \subseteq \{1, \dots, H\}\times\{1, \dots, W\}$.
    \item \textbf{Distributed patches:} Perturbations are divided across multiple small regions $\mathcal{P}_D = \{p_i\}_{i=1}^{n}$ spread throughout the image, specifically crafted to evade detection by distributing the adversarial signal, as introduced by DorPatch~\cite{he2024dorpatch}.
\end{itemize}

Given a classifier $\mathcal{F}$, original image $\mathbf{x}$, and true label $y$, the attacker's objective is to construct $\mathbf{x}_{\text{adv}}$ such that:
\begin{equation}
    \mathcal{F}(\mathbf{x}_{\text{adv}}) \ne y.
\end{equation}

\subsubsection{Attacker Capabilities}
We assume that the attacker generally has full access to the classifier's gradients and parameters, enabling effective optimization of both localized and distributed patches. Additionally, in stronger adaptive (white-box) settings explicitly evaluated in Section~\ref{sec:results}, the attacker gains knowledge of SuperPure’s super-resolution model parameters, increasing their capacity to craft robust patches. For PatchCleanser~\cite{xiang2022patchcleanser}, direct gradient-based (white-box) attacks are infeasible due to its non-differentiable masking process; thus, we specifically evaluate PatchCleanser against distributed adversarial patches crafted using DorPatch. All patch attacks adhere to realistic size constraints (patch budgets) appropriate for physical deployment scenarios, such as printed adversarial stickers.

\subsection{Defense Objective}
The defender aims to neutralize adversarial patches—both localized and distributed—while preserving accurate classification on unaltered images. To achieve broad applicability, our proposed defense method, \sys{}, meets the following key criteria:

\begin{description}
    \item[Clean Robustness.] Given a clean image-label pair $(\mathbf{x}, y)$ and a classifier $\mathcal{F}$, the defended input maintains accurate classification:
    \[
    \mathcal{F}(\sys(\mathbf{x})) = y.
    \]

    \item[Adversarial Robustness.] Given an adversarial input $\mathbf{x}_{\text{adv}}$ such that $\mathcal{F}(\mathbf{x}_{\text{adv}})\ne y$, the defended method restores the correct classification:
    \[
    \mathcal{F}(\sys(\mathbf{x}_{\text{adv}})) = y.
    \]

    \item[Model Independence.] \sys{} operates independently of the underlying classifier, requiring no modifications to classifier architecture or retraining.

    \item[Scalability and Efficiency.] \sys{} executes efficiently, suitable for real-world deployment scenarios involving high-resolution images and strict resource constraints.
\end{description}

%% file: Sections/Method.tex

In this section, we propose \sys, a novel defense strategy against adversarial patch attacks that combines a downsampling and upsampling process with a pixel-by-pixel comparison to remove both singular and distributed adversarial patches. The details of our method are presented in Algorithm \ref{alg:forward_method}.

Briefly, our algorithm first downsamples an image, $x_{adv}$, by a factor of $s$ (=4 in our setup). The downsampled image, $x_{down}$ is then fed into an upsampling method to generate a new image, $x_{up}$. The upsampled image has the same size as the original (i.e., $|x_{up}| = |x_{adv}|$). While various upsampling/image generation mechanisms exist, our experiments reveal that a GAN-based super-resolution balances accuracy, robustness, and latency. It further introduces non-linearity that helps remove the patches while retaining the information about the original image.  The new image, $x_{up}$, is compared, pixel-wise, with $x_{adv}$ using a threshold, $\lambda$ (see lines 17-22). In Section~\ref{sec:ablations}, we show how this threshold should be set. This process (downsampling, upsampling, and masking) repeats multiple times until the number of masked pixels becomes smaller than a threshold, $\epsilon$. We also report how to find this threshold in Section~\ref{sec:ablations}. The final step involves first upsampling and then downsampling the image. No masking is applied during this step (lines 11-14). The purpose of this final step is to \textit{(a)} enhance the image quality for improved accuracy and \textit{(b)} eliminate low-noise distributed patches.

We explain the theory and details of the downsampling process in \cref{sec:downsamp} and then discuss the details of the super-resolution and masking algorithm in \cref{sec:upcomp}. In \cref{sec:stoppingcond}, we outline how the method is iterated and specify the stopping condition for these iterations, and in \cref{sec:enhancement}, we describe the enhancement step. Lastly, we explain the optimizations proposed in our design to reduce the computation complexity in \cref{sec:opt}. 


\RestyleAlgo{ruled}
\LinesNumbered
\SetAlgoNoLine

\SetKwProg{Procedure}{Procedure}{}{}

\begin{algorithm}[t]
\caption{Forward Method for \sys}
\label{alg:forward_method}
\SetKwFunction{FMain}{GetDiff}
\SetKwFunction{Sys}{\sys}
\SetKwFunction{down}{DOWN}
\SetKwProg{Fn}{Procedure}{:}{}
\KwIn{Input image $\mathbf{x}_\mathrm{adv}$, max iteration count $\mathcal{K}$, masking threshold $\lambda$, stopping criterion $\epsilon$, enhance flag $\mathcal{E}$, downsampling function \down, pretrained superresolution model $G$}
\KwOut{Processed image $\mathbf{x}$}

\Fn{\Sys{$\mathbf{x}_\mathrm{adv}, \mathcal{K}, \lambda, \epsilon, \mathcal{E}$}}{
    \For{$k \leftarrow 0$ \KwTo $\mathcal{K}$}{
        $\mathbf{x}_\text{down} \leftarrow$ \down{$\mathbf{x}_\mathrm{adv}, 4$}\;
        $\mathbf{x}_\text{up} \leftarrow G(\mathbf{x}_\text{down}, 4)$\;
        $\mathbf{x}_\mathrm{adv}, c \leftarrow$ \FMain{$\mathbf{x}_\mathrm{adv}, \mathbf{x}_\text{up}, \lambda$}\;
    
        \If{c $< \epsilon$}{
            \textbf{break}\;
        }
    }
    
    $\mathbf{x} \leftarrow \mathbf{x}_\mathrm{adv}$\;
    \If{$\mathcal{E}$}{
        $\mathbf{x} \leftarrow G(\mathbf{x}, 2)$\;
        $\mathbf{x} \leftarrow$ \down{$\mathbf{x},2$}\;
    }
    
    \textbf{return} $\mathbf{x}$\;\
}

\Fn{\FMain{$\mathbf{x}_\mathrm{adv}, \mathbf{x}_\text{up}, \lambda$}}{
    $\mathbf{d} \leftarrow \mathcal{L}_2(\mathbf{x}_\mathrm{adv}, \mathbf{x}_\text{up})$\; 
    $\mathbf{m} \leftarrow \mathbf{d} > \lambda$\;
    $\mathbf{x}_\mathrm{adv} \leftarrow \mathbf{x}_\mathrm{adv} \odot \mathbf{m}$\;
  
    c $\leftarrow \sum_{i,j}{\mathbf{m}(i,j)}$\;
    \textbf{return} $\mathbf{x}_\mathrm{adv}$, c\;
}
\end{algorithm}

\subsection{Downsampling Step} \label{sec:downsamp} 

Downsampling an image reduces its spatial resolution, leading to the loss of high-frequency information; this disproportionately affects adversarial patches in comparison to the essential image content \cite{xu2018feature, shannon1949communication}. This is due to adversarial patches being heavily reliant on precise, high-frequency perturbations \cite{brown2017adversarial}. For downsampling, we utilize \textit{bilinear interpolation}, which replaces every $n \times n$ window by its average pixel value when downsampling by a factor of $n$ \cite{youssef1998analysis}. This averaging process smooths out the detailed perturbations in the patch, causing the high-frequency information needed for the patch’s effectiveness to be lost. In contrast, natural images typically display redundancy and correlation among neighboring pixels \cite{wang2004image, torralba2003statistics}. Important features and structures are often replicated across the image, making essential information less susceptible to significant degradation. 

Formally, for an image of size $H \times W$, downsampling can be modeled as a function $D_s: \mathbb{R}^{H \times W \times C} \rightarrow \mathbb{R}^{h \times w \times C}$, where $s$ is the scaling factor ($s > 1$), $h = H / s$, and $w = W / s$. Applying downsampling to the adversarial image yields: 
\begin{equation}
\mathbf{x}^\prime_{\mathrm{adv}} = D_s(\mathbf{x}_{\mathrm{adv}}) = D_s(\mathbf{x} + \boldsymbol{\delta}) = D_s(\mathbf{x}) + D_s(\boldsymbol{\delta}).
\end{equation}

Then, the energy of the adversarial patch after downsampling can be approximated as:
\begin{equation}
\|D_s(\boldsymbol{\delta})\|_2^2 \approx \frac{1}{s^2} \|\boldsymbol{\delta}\|_2^2,
\end{equation}
indicating a substantial decrease in the perturbation's magnitude due to spatial averaging.

This theoretical insight aligns with empirical evidence. Pixels within adversarial patches consistently exhibit significantly higher reconstruction errors compared to non-patch pixels; in our experiments, patch regions showed approximately $8\times$ higher errors (mean squared error of $0.6054$ vs. $0.0829$, as shown in Figure~\ref{fig:patch_recon_error} in Appendix). Additionally, prior work by Guo et al.~\cite{guo2018countering} similarly demonstrated that downsampling (e.g., via JPEG compression) effectively reduces adversarial perturbation effectiveness while preserving high accuracy on clean images. Nonetheless, simple downsampling alone remains vulnerable to adaptive attacks, underscoring the necessity of our proposed iterative and non-linear GAN-based upsampling process.


\begin{figure*}[!htp]
         \centering
         \includegraphics[width=1.4\columnwidth]{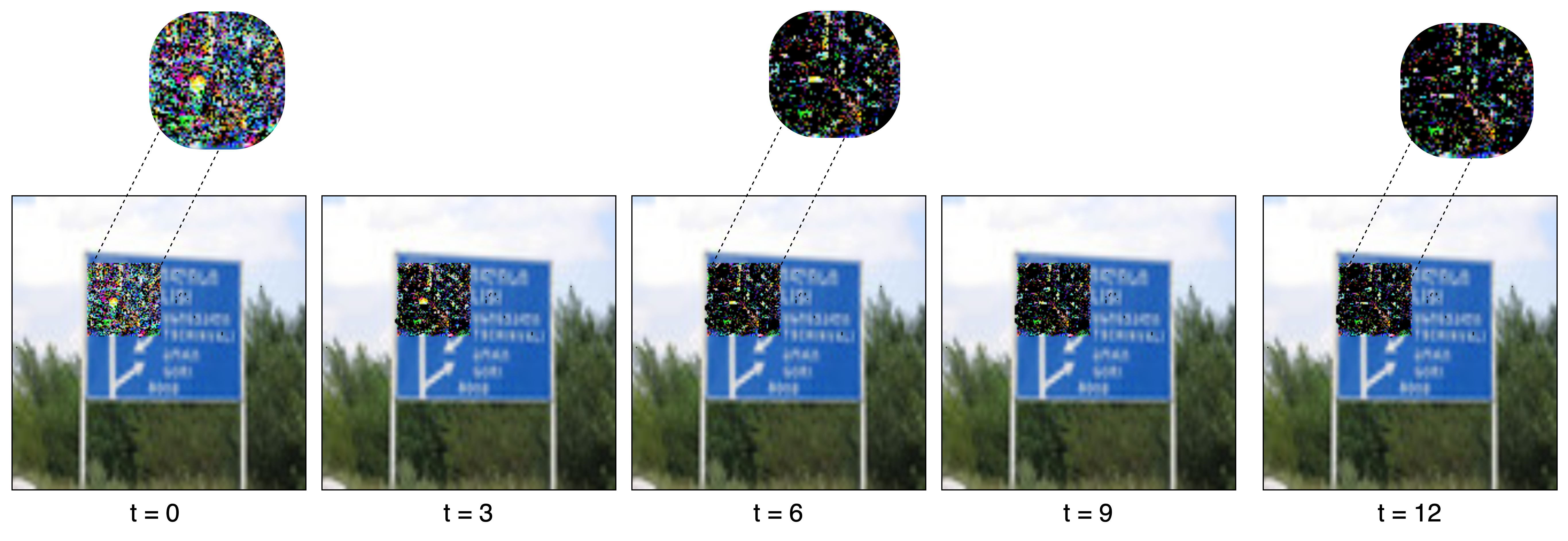}
         \caption{Shows the masking process of \ours across multiple time steps.} 
         \label{fig:timesteps}
\end{figure*}

\subsection{Upsampling and Masking} \label{sec:upcomp}
While downsampling degrades adversarial patch regions more significantly than natural non-patch areas, our initial analysis showed that applying naive down and upsampling is not sufficient to fully remove distributed patches and stronger adaptive attacks. Instead, a more powerful transformation is needed. The key idea is to apply a transformation that \textit{\textbf{disproportionally} affects adversarial regions over benign regions}. 


Based on this insight, we propose utilizing a super-resolution model for upsampling. While there are various super-resolution models available, we have chosen Real-ESRGAN \cite{wang2021real} as our primary model. Real-ESRGAN is a GAN-based model designed and trained to reconstruct high-quality images from low-resolution inputs; we use a pre-trained model provided by the authors, with no further finetuning on our datasets.

There are two reasons why the GAN struggles to reconstruct patch regions in comparison to benign non-patch regions.

First, for a pixel \( p_{i,j} \) within the adversarial patch, the surrounding pixels \( p \in B((i,j), \tau) \) are typically uncorrelated with \( p_{i,j} \) and are also heavily degraded, leaving the GAN $G$ with insufficient information for precise reconstruction. On the other hand, natural non-patch regions have more structural coherence and redundancy, leaving enough information for reconstruction even after downsampling. 

Second, since the GAN is trained on a dataset of natural images, it is optimized to generate outputs that align with the natural image distribution. However, adversarial patches deviate from this distribution, resulting in a greater pixel variation in these regions in comparison to the rest of the image. Let \( {p_{i,j}}_{a} \) denote an adversarial pixel and \( {p_{i,j}}_{c} \) a clean pixel inside an image $\pi$.
This relationship can be expressed through the following inequality:
\begin{equation}
  \mathbb{E}( |{p_{i,j}}_{a} - G(D_s(\pi))_{i,j}|) \gg  \mathbb{E}( |{p_{i,j}}_{c} - G(D_s(\pi))_{i,j}|).
  \label{eq:probs}
\end{equation}
 

This disparity results in \textit{greater pixel-wise differences} between the original image and the reconstructed output in patch regions compared to non-patch regions. Therefore, by comparing the original adversarial image $\mathbf{x}_{\mathrm{adv}}$ to the upsampled image $\mathbf{x}_{\mathrm{up}}$ on a pixel-by-pixel basis, we can mask adversarial regions by identifying where the reconstruction error exceeds a pre-defined threshold $\lambda$. Specifically, a pixel-wise comparison computes the $\mathcal{L}_2$-distance between corresponding pixels in $\mathbf{x}_{\mathrm{adv}}$ and $\mathbf{x}_{\mathrm{up}}$. A binary mask $\mathbf{m}$ is then generated, where:
$$
\mathbf{m}(i, j)= \begin{cases}1, & \text { if }\left\|\mathbf{x}_{\mathrm{adv}}(i, j)-\mathbf{x}_{\mathrm{up}}(i, j)\right\|_2>\lambda, \\ 0, & \text { otherwise. }\end{cases}
$$

The binary mask $\mathbf{m}$ is then overlaid on the adversarial image $\mathbf{x}_{\mathrm{adv}}$, effectively masking the regions that exhibit high reconstruction error and are likely to be adversarial patches.



\subsection{Iteration and Stopping Condition} \label{sec:stoppingcond}

\begin{figure}[t]
\centering
\begin{tikzpicture}
  \begin{axis}[
    width=0.4\textwidth,
    height=0.25\textwidth,
    xlabel={Step ($t$)},
    ylabel={Total Adv. Pixels Masked (\%)},
    ymin=40, ymax=100, xmin=1, xmax=10,
    xtick={1, 2, 3, 4, 5, 6,  7, 8, 9, 10},
    ytick={40, 50, 60, 70, 80, 90, 100},
    axis y line*=left,
    legend style={at={(0.5,1.3)}, anchor=north, legend columns=2, font=\small},
    legend entries={Total Masked Pixels, New Pixels Masked}
  ]
    \addplot[color=orange, mark=*, line width=1.5pt] coordinates {
(1, 49.97704349)
(2, 68.90292191)
(3, 76.98033271)
(4, 80.17760319)
(5, 81.60394317)
(6, 82.34486746)
(7, 82.74824607)
(8, 82.97173799)
(9, 83.10777655)
(10, 83.19863087)
    };

    \addlegendimage{mark=triangle*, color=purple, line width=1.5pt}
  \end{axis}

  \begin{axis}[
    width=0.4\textwidth,
    height=0.25\textwidth,
    xmin=1, xmax=10, ymin=0, ymax=2100,
    ylabel={New Pixels Masked},
    ytick={2100, 1800, 1500, 1100, 800, 500, 200, 0},
    xtick=\empty,
    axis y line*=right,
  ]
    \addplot[color=purple, mark=triangle*, line width=1.5pt] coordinates {
        (1, 2047.059701)
        (2, 775.2039801)
        (3, 330.8507463)
        (4, 130.960199)
        (5, 58.42288557)
        (6, 30.34825871)
        (7, 16.52238806)
        (8, 9.154228856)
        (9, 5.572139303)
        (10, 3.721393035)
    };
  \end{axis}
\end{tikzpicture}
\caption{Total and new adversarial pixels masked (ImageNet, Patch Size 64, Threshold=0.7)}
\label{fig:masking}
\end{figure}
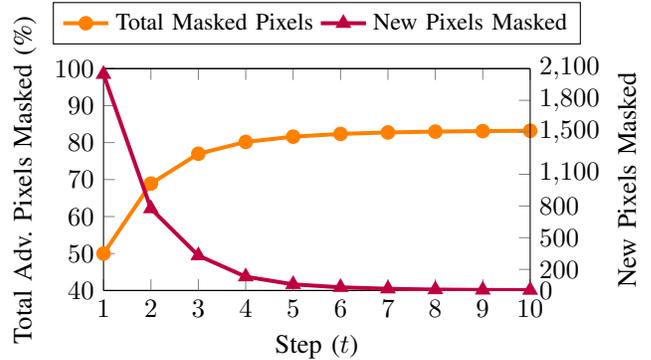

The inequality from \cref{eq:probs} suggests that with an appropriate choice of \(\lambda\) we can effectively mask adversarial pixels while preserving clean ones. However, a single iteration may not suffice to identify most adversarial pixels beyond the threshold, as a GAN can utilize local context. As a result, we use multiple iterations and observe that as the number of iterations increases, our model converges and progressively masks out adversarial pixels. This is demonstrated empirically in Figure \ref{fig:masking} as we can see the total masked pixels and total new pixels converge and steady as the number of iterations increases. The progressive masking can also be seen in Figure \ref{fig:timesteps}.

Based on the insight above, the processes presented in the previous two sections are repeated to ensure that the patch is as completely masked as possible. At each iteration, the adversarial image $\mathbf{x}_{\mathrm{adv}}$ is updated by overlaying the binary mask $\mathbf{m}$, and the new masked image becomes the input for the next iteration. This allows the adversarial regions to be gradually refined and more effectively suppressed with each cycle. As presented in \cref{fig:timesteps}, with each iteration, a greater percentage of the patch regions in the image is masked. The majority of the masking is done in earlier iterations; the number of newly masked pixels steadily decreases until the stopping condition is met. 

The stopping condition is based on the fraction of pixels that have been newly masked at a given iteration. Specifically, the iteration continues until the percentage of the newly masked pixels at iteration $t$ is below a threshold $\epsilon$:
$$
\frac{\sum^{W,H}_{i,j}{\mathbf{m_t}(i,j)}}{W\times H}<\epsilon,
$$
where $W$ and $H$ are the width and height of the image. Once the stopping condition is met, perceptible adversarial patches are mostly eliminated.



\subsection{Removing Small Distributed Perturbations} \label{sec:enhancement}

While the aforementioned iterative masking process is effective at removing perceptible adversarial patches, it struggles with smaller, distributed perturbations, such as those presented in Dorpatch \cite{he2024dorpatch}. These perturbations are subtler and often resemble natural variations within the image. Because our process relies on the assumption that the reconstruction of adversarial patches will differ significantly from that of non-patch natural image regions, this limitation of the masking process is expected.

A simple solution to mitigating these attacks involves upsampling the image via a GAN-based super-resolution model and then downsampling to the original size.

This upsampling step via super-resolution generates additional high-frequency details, while the subsequent downsampling removes that added information. The approach can be compared to DiffPure \cite{nie2022diffusion}, where random noise is added to an image before denoising it to remove any adversarial noise process. In our case, upsampling acts as the diffusion step, and downsampling serves as the denoising step, eliminating the small adversarial perturbations that are embedded within the image. Using Equation \ref{eq:probs} and its assumptions, we observe that the generated pixel is more likely to resemble the natural distribution, excluding adversarial noise. The averaging process during downsampling further reduces this noise by smoothing out pixel-level variations, while the new clean pixel generated by the GAN helps to replace adversarial components. Together, these steps effectively reduce the overall noise, acting as a surrogate for the denoising step in diffusion models.

This process is performed after the initial iterative masking step, ensuring that any remaining subtle adversarial perturbations are addressed. We chose to upsample the image before downsampling (instead of vice versa) because starting with downsampling discards significant information in the image, leading to a loss of essential details and poor classifier performance. The final processed image can then be used for downstream tasks, ensuring that the adversarial manipulations no longer have a significant effect on model performance. We use \sysp as a shorthand for our method with this enhancement in subsequent sections.

\subsection{Computation Overhead Optimizations} \label{sec:opt}
While our main goal is to improve the system's robustness against singular (localized) and distributed patch attacks, we've made a series of design decisions in order to reduce \sys's computation overhead. In this part, we briefly discuss them and explain other alternatives. 

\vspace{2pt}
\noindent \textbf{Down vs. Upsampling.} While \sys utilizes a down-sampling step for patch removal, our initial analysis showed that a similar upsampling (using super-resolution) and then down-sampling could result in similar robustness. 
The main reason is that although the steps are swapped, the joint processing pipeline remains (relatively) the same hence similar theoretical analysis could be applied (i.e., super-resolution is harder for patch areas and down-sampling helps with filtering). Furthermore, as shown in Section~\ref{sec:enhancement}, the up-down sampling option has an advantage over down-up since it can remove the distributed noises more effectively. 

The key difference between the two options, however, relies on computation efficiency.  In more detail, the up-down sequence increases the computation overhead because the more costly operation (super-resolution) needs to be performed on the larger image sizes (i.e., super-resolution on the original input size vs. super-resolution on the downsampled image). As a result, for the iterative masking part, we chose the down-up method instead of the up-down method to favor efficiency. However, to achieve enhancement, we add one final layer of up-down sampling. This way, accuracy-robustness-latency can be jointly optimized. 

\vspace{2pt}
\noindent \textbf{Super-Resolution Model.} An important component in our design is the super-resolution model. Among various options, we opt for a GAN-based model since we found that it can achieve the right balance between robustness and complexity. The alternative option is using a more sophisticated model, e.g., a diffusion-based system~\cite{nie2022diffusion}. While we expect that such a model would perform better (in terms of robustness and accuracy), it incurs orders of magnitude higher overhead (latency, memory, etc.). 

Alternatively, the other extreme is using a much simpler upsampling strategy to further reduce the complexity. While we considered this, our initial analysis showed that simpler models are significantly more vulnerable to low-noise distributed attacks. Even worse, they are far more vulnerable to adaptive white-box attacks where an adversary creates patches that are resistant to up/downsampling. In Section~\ref{sec:ablations}, we study our model's robustness against white-box attacks and show that \sys retains decent robustness even in the presence of an adaptive attack. 

\vspace{2pt}
\noindent \textbf{Stop Condition.} Another important factor in optimizing end-to-end latency is the stop condition. There is a tradeoff between the number of iterations and end-to-end latency. On one hand, more iterations are necessary to enhance robustness (see Figure~\ref{fig:timesteps}). On the other hand, fewer iterations result in lower latency. We address this balance by implementing a dynamic stop condition method (see lines 6-7 in Algorithm 1) based on a user-defined parameter ($\epsilon$). In Section~\ref{sec:ablations}, we examine the impact of different thresholds on robustness. Overall, our results demonstrate that \sys can achieve high robustness without requiring an excessively large number of iterations (fewer than ten on average), enabling it to attain both efficiency and robustness simultaneously.

\begin{figure*}[th!]
         \centering
         \includegraphics[width=1.4\columnwidth]{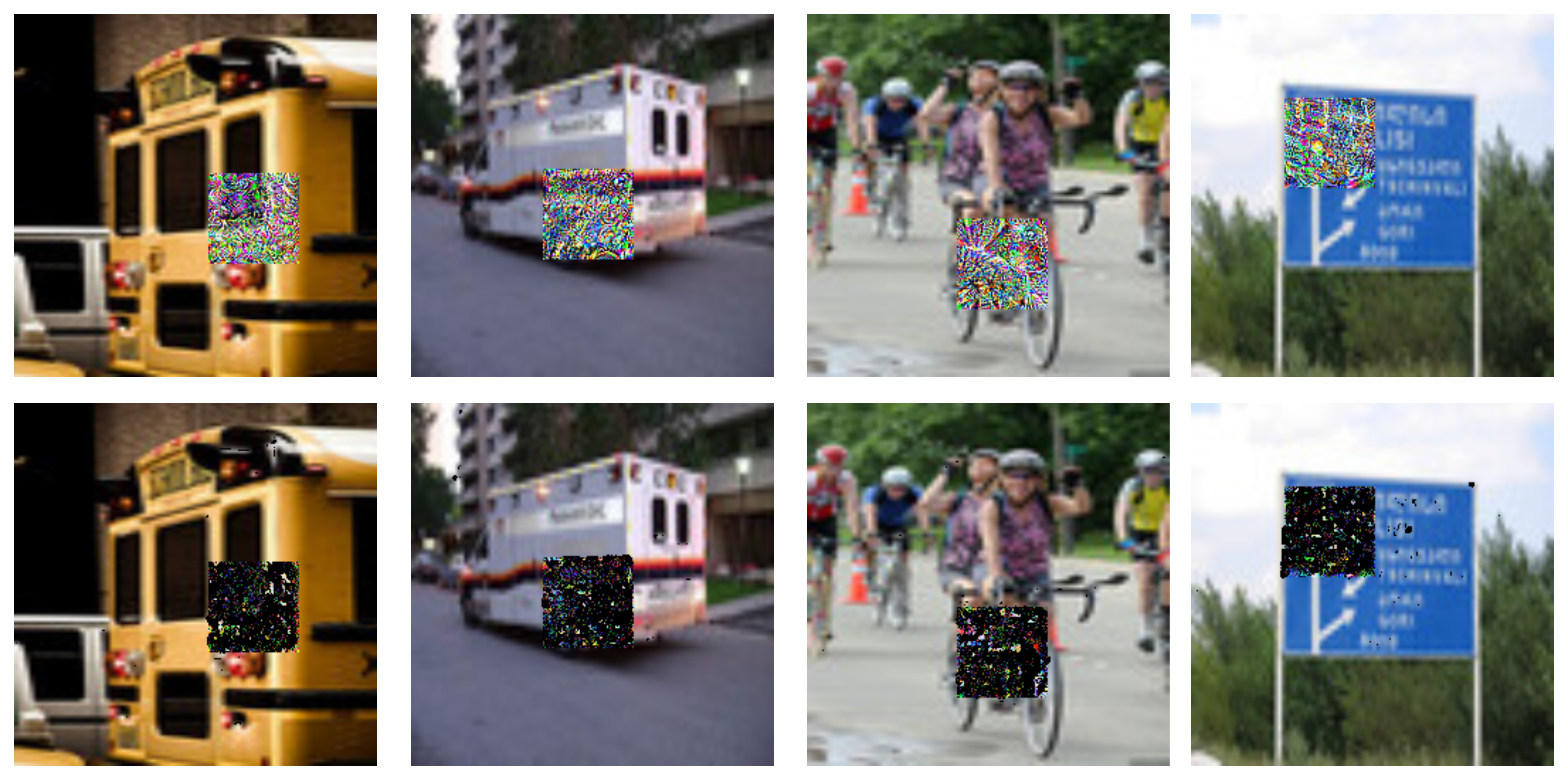}
         \caption{Quality results for \sys before (top) and after purification (down).} 
         \label{fig:orgvpurified}
\end{figure*}

%% file: Sections/new_results_5.tex

In this section, we provide a comprehensive evaluation of \sys. We describe the experimental setup in \cref{sec:expsetup}, followed by a presentation of the robustness analysis against various patch attacks in \cref{sec:adveval}. We also report the computation overhead results (latency and memory usage) in Section~\ref{sec:latency}. 

Specifically, the goal of our analysis is to answer the following research questions:
\begin{itemize}
    \item Q1: Compared to state-of-the-art, how robust \sys is against singular (localized) patch attacks? 
    \item Q2: Is \sys robust against distributed patch attacks (as opposed to state-of-the-art)?
    \item Q3: Is \sys robust against white-box attacks?
    \item Q4: Compared to prior methods, what is the computation complexity of \sys?
\end{itemize}

In addition to answering these questions, we provide an extensive ablation study in Section~\ref{sec:ablations} to further analyze the important factors in jointly optimizing robustness-accuracy-latency. 



     

\subsection{Experimental Setup} \label{sec:expsetup}

In this section, we detail the experimental setup used to evaluate the effectiveness of our proposed method against adversarial patch attacks. We describe the datasets, classifiers, super-resolution setup, attack configurations, and evaluation metrics.

\subsubsection{Datasets}

Similar to prior work~\cite{xiang2022patchcleanser}, we conduct our experiments using the ImageNet dataset \cite{deng2009imagenet}, specifically utilizing a subset of the validation set to assess the performance of our method. From the validation set, we select five images per class, resulting in a total of 5,000 images across the 1,000 classes. This subset provides a diverse and representative sample for evaluating the robustness of our approach.

\subsubsection{Classifiers}

To demonstrate that our method is truly classifier-agnostic, we evaluate it on three \textit{architecturally distinct} network families: EfficientNet-B0~\cite{tan2019efficientnet}, ResNet-152 v2~\cite{he2016identity}, and ViT-B/16~\cite{dosovitskiy2021an}.

\textbf{EfficientNet-B0} is part of the EfficientNet family, which utilizes compound scaling to balance network depth, width, and resolution. We use a PyTorch-provided checkpoint trained on ImageNet~\cite{paszke2019pytorch}.

\textbf{ResNet-152 v2} is a deep residual network with 152 layers, incorporating identity mappings to facilitate the training of very deep architectures. We use a PyTorch-provided checkpoint trained on ImageNet~\cite{paszke2019pytorch}.

\textbf{ViT-B/16} is a Vision Transformer model that tokenizes an input image into $16\times16$ patches and processes them via a Transformer-based architecture to capture global context. We use a PyTorch-provided checkpoint trained on ImageNet~\cite{paszke2019pytorch}.

These three models are notably diverse in design: EfficientNet focuses on compound scaling, ResNet employs deep residual blocks, and ViT adopts a patch-based Transformer architecture. By testing on these fundamentally different paradigms, we thoroughly assess the robustness and generality of our proposed defense across a wide spectrum of network architectures.

\subsubsection{\sys and Super-Resolution Model Setup}

Our method leverages super-resolution techniques to mitigate the impact of adversarial patches. Specifically, we use the Real-ESRGAN models \cite{wang2021real} for image upsampling, which serve to diminish the adversarial perturbations introduced by the attacks. We choose Real-ESRGAN due to \textit{(a)} its minimal domain overlap with ImageNet (it is not trained on ImageNet), \textit{(b)} the availability of publicly released checkpoints for reproducibility, and \textit{(c)} its GAN-based speed advantage over diffusion-based methods—crucial when it must be called multiple times in our pipeline.

We employ two versions of Real-ESRGAN; one that upsamples images by a factor of four, and another by a factor of two. The choice of these numbers is based on our preliminary analysis. We also considered other factors, but ultimately observed that four and two achieve the best balance between robustness and latency. 

\begin{table*}[t]
\centering
\caption{Robustness of different defense methods against single (localized) adversarial patch attacks on ViT, EfficientNet, and ResNet classifiers.}
\label{tab:superpure_robustness}
\begin{tabular}{@{}llcccccc@{}}
\toprule
\multirow{2}{*}{\textbf{Model}} & \multirow{2}{*}{\textbf{Defense Method}} & \multicolumn{6}{c}{\textbf{Patch Size}} \\
\cmidrule(lr){3-8}
& & 0 (no attack) & 16$\times$16 & 32$\times$32 & 48$\times$48 & 64$\times$64 & 96$\times$96 \\

\midrule

\multirow{6}{*}{ViT} 
    & No Defense    & 74.84 & 38.02 & 4.32 & 00.50 & 0.16 & 0.00 \\
    & PatchCleanser~\cite{xiang2022patchcleanser} & 72.10 & 54.33 & 44.21 & 35.3 & 30.74 & 20.72 \\
    & PAD ~\cite{jing2024pad}                 & 44.76 & 46.58 & 47.04 & 46.36 & 45.62 & 41.64 \\
    & \sys                 & 74.96 & 74.30 & 73.66 & 72.86 & 70.36 & 62.76 \\
    & \sysp       & \textbf{82.98} & \textbf{80.70} & \textbf{77.82} & \textbf{77.52} & \textbf{74.66} & \textbf{65.9} \\
\midrule
\multirow{6}{*}{EfficientNet}  
    & No Defense    & 60.76 & 30.82 & 5.12 & 0.82 & 0.20 & 0.02\\
    & PatchCleanser~\cite{xiang2022patchcleanser} & 57.98 & 43.60 & 38.46 & 27.46 & 22.66 & 10.92 \\
    & PAD~\cite{jing2024pad}                  & 34.70 & 35.30 & 34.70 & 33.94 & 32.72 & 26.02 \\
    & \sys                & 61.10 & 59.76 & 58.34 & \textbf{55.98} & \textbf{52.42} & \textbf{41.86} \\
    & \sysp       & \textbf{69.08} & \textbf{63.54} & \textbf{60.48} & 54.12 & 46.72 & 28.22 \\
\midrule

\multirow{6}{*}{ResNet} 
    & No Defense    & 71.70 & 45.10 & 24.52 & 14.28 & 4.38 & 0.10 \\
    & PatchCleanser~\cite{xiang2022patchcleanser} & 68.98 & 56.72 & 51.64 & 41.28 & 33.82 & 19.66 \\
    & PAD ~\cite{jing2024pad}                 & 48.19 & 50.10 & 50.28 & 49.38 & 45.04 & 43.30 \\
    & \sys                 & 71.20 & 70.48 & 70.10 & 68.52 & 66.20 & \textbf{58.18} \\
    & \sysp       & \textbf{79.86} & \textbf{76.74} & \textbf{76.30} & \textbf{74.20} & \textbf{70.64} & 57.84 \\
\bottomrule
\end{tabular}
\end{table*}

For both models, we use the checkpoints provided by the original authors, which were trained on the DIV2K \cite{agustsson2017ntire}, Flickr2K \cite{timofte2017ntire}, and OutdoorSceneTraining (OST) dataset \cite{wang2018recovering}. Note that ImageNet, our evaluation dataset, was \underline{not} included in the Real-ESRGAN training set. We believe that fine-tuning on ImageNet could further improve our results. However, our following results show that \sys can achieve excellent performance even without fine-tuning. 

Unless stated otherwise, for all our experiments, we set $\lambda=.7$ and $\epsilon=4$ (pixels). 

\subsubsection{Adversarial Attacks}

To evaluate the robustness of our method, we generate adversarial patches using the Masked Projected Gradient Descent (Masked PGD) method. Following the approach used in PatchGuard \cite{xiang2021patchguard} and PatchCleanser~\cite{xiang2022patchcleanser}, we use the following attack configuration:

\begin{itemize} 
    \item \textbf{Maximum Perturbation ($\epsilon$)}: Set to $1$ (with pixel values normalized to the $[0,1]$ range). 
    \item \textbf{Number of Iterations}: $100$ iterations. 
    \item \textbf{Step Size}: $0.05$ per iteration. 
    \item \textbf{Random Start}: Each attack began from a random point within the allowed perturbation range. 
    \item \textbf{Optimization Strategy}: Loss function evaluated every $10$ iterations to select the sample that maximized the loss.
    \item \textbf{Patch Location}: Randomized for each image to simulate unpredictable attack scenarios. 
    \item \textbf{Patch Sizes}: Patches of sizes $16\times 16$, $32\times 32$, $48\times 48$, $64\times 64$, and $96\times 96$ pixels (approximately $0.5\%, 2\%, 4.6\%, 8.2\%,$ and $18.4\%$ of the average size of ImageNet \cite{deng2009imagenet} images respectively).
\end{itemize}

In addition to singular localized patches, we also consider distributed patches. Specifically, we create new patches using the method proposed by DorPatch~\cite{he2024dorpatch}. This attack uses a patch budget of 12\%, a density of 0.1\%, and a maximum of $5,000$ iterations. It evaluates the ability of our method to handle sparse yet effective perturbations designed to evade detection.

\subsubsection{Evaluation Metrics}

We use clean accuracy (top-1) and robust accuracy as our primary evaluation metrics. Clean accuracy refers to the proportion of clean test images that our defended model correctly classifies. Robust accuracy is similarly defined as the proportion of adversarial test images accurately classified by our model. Additionally, we measure per-example inference time (s/img) to assess computational overhead.

We also report the defense performance of two state-of-the-art methods: PAD \cite{jing2024pad} and PatchCleanser \cite{xiang2022patchcleanser}, for comparison. We use the optimal defense settings stated in their respective papers; note that since PatchCleanser is dependent on patch size, we tested multiple configurations (window sizes) and chose the best results to report. Figure~\ref{fig:orgvpurified} presents the quality results.

\subsubsection{Hardware Setup for Latency Measurement} All latency measurements are conducted using an NVIDIA RTX 4090 GPU with 24 GB of memory to ensure precise and consistent evaluation of computational performance. We use PyTorch version 1.12. Our code and results will be \textbf{\textit{open-source}}.

\subsection{Robustness Results} \label{sec:adveval}
In this subsection, we present robustness results for our defense against singular black-box and white-box patches of varying sizes. We also present results against distributed patches.

\subsubsection{Single Patch Attacks} \label{sec:singeval}
\cref{tab:superpure_robustness} compares the robustness of our defense method against PatchCleanser \cite{xiang2022patchcleanser} and PAD \cite{jing2024pad} for clean (patch size 0) and singular adversarial patch samples on EfficientNet and ResNet models. 

We denote our approach without the last enhancement step as \sys and with enhancement as \sysp. As shown in \cref{tab:superpure_robustness}, both \sys and \sysp demonstrate significant robustness improvements over other defense methods under varying patch attack sizes. 

On the smallest patches of size $16\times 16$, \sysp achieves 63.54\% and 76.74\% Top-1 accuracy on EfficientNet and ResNet, respectively, compared to 30.82\% and 45.10\% for the baseline with no defense. For a patch size of $48\times 48$, \sysp achieves 54.12\% accuracy on EfficientNet and 74.20\% on ResNet, while the highest-performing baseline defense (PAD) only reaches 33.94\% and 49.38\% on the respective models. Even with the largest patch size tested ($96\times 96$), \sysp achieves 28.22\% Top-1 accuracy on EfficientNet and 57.84\% on ResNet. 
This result is significantly higher than the no-defense baseline (close to 0\% for both classifiers) and surpasses PatchCleanser and PAD, confirming our method’s robustness against severe adversarial perturbations. Similar results can be observed for ViT, where \sys and \sysp consistently perform better compared to prior works. 

An additional noteworthy aspect of our results is the \textit{clean accuracy} achieved by both \sys and \sysp. On clean, unaltered images (patch size 0), \sys maintains strong Top-1 accuracy, while \sysp exhibits even better performance, achieving 82.98\% on ViT, 69.08\% on EfficientNet, and 79.86\% on ResNet. These results underscore that our defense methods not only provide mitigations for attacks but also retain high accuracy for benign situations.

For \sysp, the clean accuracy is higher than the baseline configuration with no defense and no attack by around 8\% for all models (74.84\% for ViT, 60.76\% for EfficientNet, and 71.7\% for ResNet). Specifically for the ResNet classifier, \sysp achieves higher accuracy on adversarial samples than the baseline no defense on clean images up to a patch size of $48\times 48$, and shows comparable performance at $64\times 64$. This indicates that our enhancement step contributes a dual benefit: \textbf{superior robustness to adversarial patches and improved accuracy on clean inputs}. The key reason for this is due to the super-resolution-based enhancement strategy described in Section~\ref{sec:enhancement}. We further study the impact of enhancement in our ablation study (Section~\ref{sec:evalenhance}).

\subsubsection{Distributed Patch Attacks}
Unlike localized adversarial patches that occupy a single, contiguous
region, \emph{distributed} patches fragment perturbations into multiple
sub-regions, making it harder for a single masking window to neutralize
the entire patch. DorPatch~\cite{he2024dorpatch} follows this principle
by scattering its patch budget across many smaller pieces. Consequently,
each piece can use a lower noise amplitude, which both reduces visibility
and maintains high adversarial effectiveness.

\Cref{tab:dorpatch_results} compares our defense with
PatchCleanser and PAD
against DorPatch. Even though DorPatch and standard patches share the
\emph{same overall budget}, distributing that budget means partial masking
has less impact on the attack. This makes \emph{certifiable} defenses like
PatchCleanser—which relies on covering a single contiguous area—ineffective
(robustness stays at 0\% under DorPatch), whereas \sysp still defends
successfully with 59\% accuracy.

\begin{table}[b]
\centering
\caption{Effectiveness against distributed DorPatch~\cite{he2024dorpatch} attack. The enhancement step in \sysp makes our system robust.}
\label{tab:dorpatch_results}
{%
\begin{tabular}{lcc}
\toprule
\textbf{Method} & \textbf{Clean Accuracy} & \textbf{DorPatch Robustness} \\
\midrule
No Defense                 & 72\% & 0\%   \\
PatchCleanser~\cite{xiang2022patchcleanser}  & 69\% & 0\%   \\
PAD~\cite{jing2024pad}    & 48\% & 39\%   \\
\sys & 71\% & 0\% \\
\sysp                      & \textbf{80\%} & \textbf{59\%} \\
\bottomrule
\end{tabular}
}
\end{table}

While \sys cannot purify distributed patches, the \emph{enhancement} step in \sysp (see \cref{sec:enhancement})
remains effective at removing these scattered perturbations, allowing our
defense to handle sophisticated multi-patch threats that circumvent
single-window defenses like PatchCleanser.

Notably, DorPatch leverages \emph{low noise amplitude} patches that
subtly alter the image, rendering a purely threshold-based masking (i.e.,
\sys -- without enhancement) ineffective: the per-pixel
differences often lie below the threshold and thus remain unmasked. In
contrast, \sysp adds an \emph{enhancement} step 
to purify small, distributed perturbations, allowing our defense to handle
DorPatch’s scattered, low-amplitude noise when other single-window or
simple threshold-based defenses fail.

\subsubsection{Adaptive White-Box Attacks}

We also evaluate our defense under an adaptive \emph{white-box} threat model, where the adversary has complete access to both our purification network (including the super-resolution module and our method) and the target classifier. This means the attacker can compute gradients through every component of our defense to craft adversarial patches tailored explicitly to our defense mechanism. 

\begin{table}[t]
\centering
\caption{Robust Accuracy (\%) under White-Box Attacks on ResNet.}
\label{tab:whitebox}
\begin{tabular}{lcc}
\toprule
\multirow{2}{*}{\textbf{Defense}} & \multicolumn{2}{c}{\textbf{Patch Size}} \\
\cmidrule(lr){2-3}
 & 48$\times$48 & 64$\times$64 \\
\midrule
Na\"ive Down\&Up (white-box) & 9.12 & 4.89 \\
PatchCleanser \cite{xiang2022patchcleanser} (non--white-box) & 41.28 & 31.28 \\
\sys/\sysp{} (ours, white-box) & \textbf{60.38} & \textbf{51.52} \\
\bottomrule
\end{tabular}
\end{table}

As shown in Table~\ref{tab:whitebox}, which compares the robust accuracy of three different defenses—\emph{Na\"ive Down\&Up}, \emph{PatchCleanser}, and our method \emph{\sysp{}}—against adversarial patches of size $48\times48$ and $64\times64$ on ResNet\footnote{We report results for two patch sizes on a single model due to the high computational cost of white-box experiments. However, based on prior findings, we believe these results are representative and likely transferable to other models and patch sizes.}, a simple ``Na\"ive Down\&Up'' defense completely fails when facing an adaptive attacker. This naive approach applies a fixed downsampling followed by upsampling in hopes of smoothing out adversarial noise. However, under a white-box setting where the attacker has the knowledge of this defense, it becomes trivial for the attacker to generate perturbations that survive such transformations. As a result, the robust accuracy drops sharply to only 9\% and 5\% for patch sizes of $48\times48$ and $64\times64$, respectively.

PatchCleanser, which is not differentiable and thus not directly applicable in a white-box setting, is evaluated here under its own original (black-box) setup. While it performs better than the naive method in that setting (41.28\% and 31.28\%), it still falls short compared to our method. In contrast, our proposed \sysp{} is evaluated in a fully adaptive white-box setting, where the attacker has complete access to the super-resolution model and can backpropagate through the entire pipeline. Despite this, our method maintains robust accuracy of 60.38\% and 51.52\%, significantly outperforming both the naive baseline and PatchCleanser—even though the latter operates under a more favorable (black-box) scenario. This strong robustness, even under white-box conditions, stems from the inherent nonlinearity and complexity of our defense pipeline. Unlike simple smoothing operations, our method first projects inputs into a more natural image manifold using a deep super-resolution network, then applies a masking and enhancement mechanism that further disrupts adversarial structures. Because the entire purification process is nonlinear and includes operations that do not preserve gradients in a predictable way, it becomes significantly harder for the attacker to generate perturbations that survive all these transformations. As a result, even with full access to our model and the ability to compute gradients through it, the adversary struggles to construct successful attacks. The resulting drop in robust accuracy compared to the black-box setting is relatively small, demonstrating that our method retains strong practical resilience—even in challenging adaptive white-box scenarios.


\subsection{Computation Overhead} \label{sec:latency}

\subsubsection{Latency} \label{sec:latency-sub}

Another major benefit of \sys/\sysp is its superior end-to-end latency compared to prior work. To highlight this, we compare \sys/\sysp to PatchCleanser and PAD. Specifically, we employ two different setups for Patch Cleanser. One setup used a larger masking window, resulting in fewer masks (16), which we refer to as \textit{PatchCleanser-Efficient}. The other setup used a smaller masking window, increasing the number of masks (49) for greater robustness. These setups are evaluated on ResNet.


\begin{table}[tbp]
    \centering
    \caption{End-to-end latency.}
    \label{tab:speed}
    \begin{tabular}{lc}
    \toprule
    \textbf{Method} & \textbf{Time (s)} \\
    \midrule
   \sys/\sysp        & \textbf{0.53 / 0.58} \\
    PatchCleanser-Efficient/ PatchCleanser \cite{xiang2022patchcleanser}  & 3.89 / 36.63 \\
    PAD                           & 8.80 \\
    \bottomrule
    \end{tabular}
\end{table}

As shown in \cref{tab:speed}, our method \sysp{} takes 0.58 seconds per image, and \sys{} takes 0.53 seconds per image, both significantly faster than even Patch Cleanser Efficient (3.89 seconds). Additionally, our approach is substantially faster than regular Patch Cleanser (36.63 seconds) and PAD (8.8 seconds). This significant reduction in latency makes our approach \textbf{\textit{more practical}} for real-world scenarios while still remaining effective against both localized and distributed physical patch attacks.

The key reason for this improvement over state-of-the-art methods lies in the design decisions detailed in Section~\ref{sec:opt}. Specifically, compared to PatchCleanser, \sysp{} requires significantly fewer steps. Likewise, compared to PAD, our per-iteration analysis is much simpler, involving only a GAN-based super-resolution and straightforward downsampling.

\subsubsection{GPU Memory Overhead} \label{sec:gpu}
\cref{tab:gpu_overhead} provides the GPU memory overhead comparison for each method, excluding the classifier. Since PatchCleanser does not rely on any external model, its overhead is effectively zero. In contrast, both our method and PAD require external models, resulting in higher GPU overhead. This trade-off, however, is justified by the improved robustness and speed offered by our approach.

\begin{table}[htbp]
\centering
\caption{GPU Memory Overhead}
\label{tab:gpu_overhead}
{%
\begin{tabular}{lc}
\toprule
\textbf{Method} & \textbf{GPU Memory Usage} \\
\midrule
PatchCleanser~\cite{xiang2022patchcleanser}      & \textbf{0}   \\
PAD~\cite{jing2024pad}       & 5290M   \\
\sys     & 1895M  \\
\sysp     & 2087M  \\
\bottomrule
\end{tabular}
}
\end{table}

\subsection{Comparison with PatchCURE}

We also compare \sys{} against PatchCURE~\cite{xiang2024patchcure}, a recent state-of-the-art extension of PatchCleanser that modifies model architectures (e.g., ViT-SRF) to improve \emph{performance} and efficiency while preserving roughly the same robustness as PatchCleanser. Its pipeline first uses an SRF (small receptive field) sub-model to extract an intermediate feature map so that only part of the features is corrupted; it then applies a secure operation that typically requires multiple calls to a large receptive field (LRF) sub-model. Because these architectural changes require partial retraining, PatchCURE is not a simple plug-and-play solution and is thus omitted from our main comparisons (\Cref{sec:adveval}). Instead, we conduct a targeted evaluation on ImageNet classification with a $32\times32$ patch, using the \emph{pretrained ViT-SRF} model from PatchCURE. For our own defense, we rely on the same pretrained ViT model that underpins our broader evaluations, ensuring consistency across all tested approaches.

\begin{table}[htbp]
\centering
\caption{Comparison with PatchCURE~\cite{xiang2024patchcure} on ViT (ImageNet, $32\times32$ patch). ``Retrain'' indicates whether the method requires classifier retraining.}
\label{tab:patchcure_comparison}
\begin{tabular}{lccc}
\toprule
\textbf{Method} & \textbf{Clean (\%)} & \textbf{Robust (\%)} &  \textbf{Retrain?} \\
\midrule
PatchCURE~\cite{xiang2024patchcure} & 72 & 41 &  Yes\\
\sys{}                      & 80 & 75 &  No\\
\sysp{}                       & \textbf{83} & \textbf{79} &  No\\
\bottomrule
\end{tabular}
\end{table}

As shown in \Cref{tab:patchcure_comparison}, our approaches \sys{} and \sys{}+ substantially outperform PatchCURE in both clean accuracy and robustness. In particular, \sys{}+ achieves an absolute gain of 38\% in robustness and 11\% in clean accuracy compared to PatchCURE. More importantly, PatchCURE still mandates partial retraining when modifying the classifier, increasing deployment complexity and overhead. By contrast, our approach is entirely plug-and-play, requiring no classifier modifications—thus highlighting the \textbf{practicality, efficiency, and scalability} of our proposed method for real-world applications.

\begin{table}[htbp]
\centering
\caption{Latency comparison on a Jetson device for ViT-based setups.}
\label{tab:patchcure_latency}
\begin{tabular}{lcc}
\toprule
\textbf{Method} & \textbf{Latency (s)} & \textbf{Repeated Classifier?} \\
\midrule
PatchCleanser~\cite{xiang2022patchcleanser} & $>$50 & Yes \\
PatchCURE~\cite{xiang2024patchcure} & 12 & Partial \\
\sys{} & \textbf{0.67} & No (single pass)\\
\sysp & 0.72 & No (single pass)\\
\bottomrule
\end{tabular}
\end{table}

Finally, \Cref{tab:patchcure_latency} compares the end-to-end latency on a Jetson device using ViT-based setups. PatchCleanser requires over 50\,s, whereas PatchCURE reduces the total inference time to about 12\,s. By contrast, our method needs only \textbf{0.67\,s}—thanks to a single-pass purification strategy that eliminates repeated classifier calls. These results confirm that \sys{} exceeds PatchCURE not only in accuracy and robustness but also in latency-critical scenarios, all while maintaining a plug-and-play design with no modifications to existing classifiers.

%% file: Sections/Ablations.tex
\subsection{Iterations to Convergence} 
\begin{figure}[h]
\centering 
\begin{tikzpicture}
  \begin{axis}[
    width=0.4\textwidth, 
    height=0.27\textwidth,
    xlabel={Patch Size},
    ylabel={Top-1 Accuracy (\%)},
    ymin=0, ymax=85, xmin=0, xmax=96,
    xtick={0,16,32,48,64,80,96},
    ytick={5,15,25,35,45,55,65,75,85},
    axis y line*=left,
    label style={font=\small},
    legend style={at={(0.5,-0.27)}, anchor=north, legend columns=2, font=\small},
    legend entries={No Defense Accuracy, \sys Accuracy, \sysp Accuracy, Average Iterations}
  ]
    \addplot[color=orange, mark=square*, line width=1.5pt] coordinates {
      (0,71.70)
      (16,45.1)
      (32,24.52)
      (48,14.28)
      (64,4.38)
      (96,0.1)
    };
    \addplot[color=purple, mark=triangle*, line width=1.5pt] coordinates {
      (0,71.20)
      (16,70.48)
      (32,70.10)
      (48,68.52)
      (64,66.20)
      (96,58.18)
    };
    \addplot[color=black, mark=*, dashed, line width=1.5pt] coordinates {
      (0,79.86)
      (16,76.74)
      (32,76.3)
      (48,74.2)
      (64,70.64)
      (96,57.84)
    };
    \addlegendimage{mark=*, color=teal, line width=1.5pt}

  \end{axis}
  
  \begin{axis}[
    width=0.4\textwidth, 
    height=0.27\textwidth,
    xmin=0, xmax=96, ymin=3, ymax=18,
    ylabel={Average Iterations},
    ytick={4,6,8,10,12,14,16,18},
    xtick=\empty,
    axis y line*=right,
    label style={font=\small},
  ]
    \addplot[color=teal, mark=*, line width=1.5pt] coordinates {
      (0,3.2488)
      (16,7.0382)
      (32,10.0142)
      (48,11.903)
      (64,13.309)
      (96,15.2452)
    };
  \end{axis}
\end{tikzpicture}
\caption{Relationship between patch size, accuracy, and average number of iterations until convergence.}
\label{fig:patch_size_iter}
\end{figure}
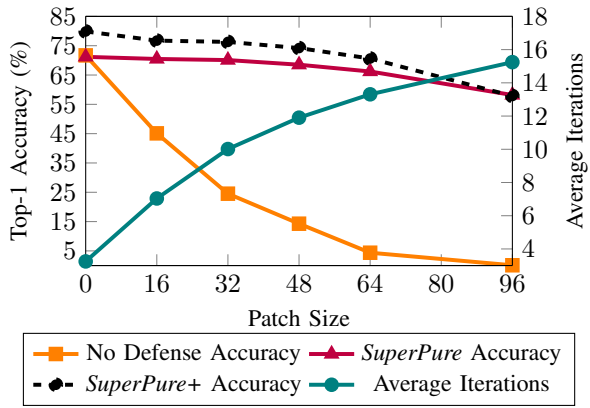

Figure \ref{fig:patch_size_iter} demonstrates the relationship between patch size, Top-1 accuracy, and the average number of iterations to convergence for  ResNet model in the contexts of \sys and \sysp.

As seen in the figure, the average number of iterations needed for convergence increases as the patch size grows, suggesting that our method is able to \textbf{adapt to different patch sizes} for computational efficiency. In scenarios where there is no patch present, \sysp only requires around 3 iterations on average before stopping. As the patch size increases, the number of iterations rises accordingly, reflecting the greater complexity of defending against larger adversarial patches. Note that the number of iterations for EfficientNet and ResNet are slightly different because patches are generated specifically for each classifier.

\subsection{Effect of Enhancement} \label{sec:evalenhance}
An integral component of our algorithm includes the option to enable or disable a feature we refer to as ``enhancement,'' i.e., the difference between \sys and \sysp. As described in \cref{sec:enhancement}, this feature involves a two-step process where the input image is first up-sampled to a higher resolution and then down-sampled back to its original dimensions. In \cref{tab:superpure_robustness}, we observe that enhancement is not only essential for defending against smaller, distributed patches but also beneficial for robustness for clean images and singular adversarial patches. 

\begin{figure}[t]
    \centering
    \includegraphics[width=.85\columnwidth]{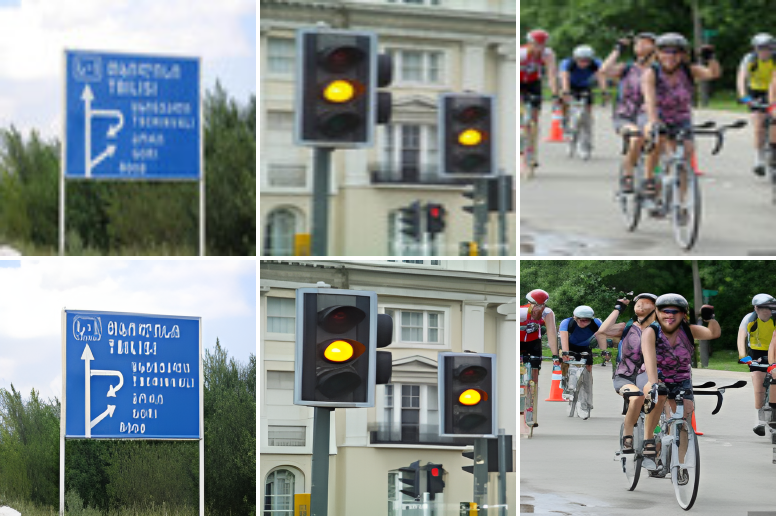}
    \caption{Clean images before (top) and after (bottom) enhancement.} 
    \label{fig:qualenhance}
\end{figure}

To better understand the role of enhancement, we conduct experiments where we apply this process, without iterative masking, to clean images. In addition to the ResNet and EfficientNet architectures, we evaluate three other classifiers: VGG-16 with batch normalization \cite{simonyan2015very}, WideResNet-50-2: \cite{zagoruyko2016wide}, and ViT-B/16 \cite{dosovitskiy2021an}. The results reported in \cref{tab:entop1_accuracy} reveal that top-1 accuracy increases by an average of approximately 10 percentage points. We can see in \cref{fig:qualenhance} that enhancement improves the visual quality of images, with clearer boundaries and more defined textures, which may aid the model in focusing on key features for classification.

\begin{table}[htbp]
\centering
\caption{The impact of enhancement on top-1 clean accuracy.}
\label{tab:entop1_accuracy}
{%
\begin{tabular}{lccc}
\toprule
\textbf{Model} & \textbf{Standard} & \textbf{Enhanced} & \textbf{Change} \\
\midrule
EfficientNet-B0\cite{tan2019efficientnet}    & 60.76\% & 70.60\% & +9.84\% \\
WideResNet-50-2\cite{zagoruyko2016wide}    & 61.00\% & 74.46\% & +13.46\% \\
VGG-16 with BN\cite{simonyan2015very}     & 47.12\% & 58.28\% & +11.16\% \\
ViT-B/16\cite{dosovitskiy2021an}           & 74.84\% & 84.82\% & +9.98\% \\
ResNet-152 V2\cite{he2016identity}      & 71.70\% & 81.52\% & +9.82\% \\
\bottomrule
\end{tabular}
}
\end{table}




\subsection{Effect of Masking Threshold}

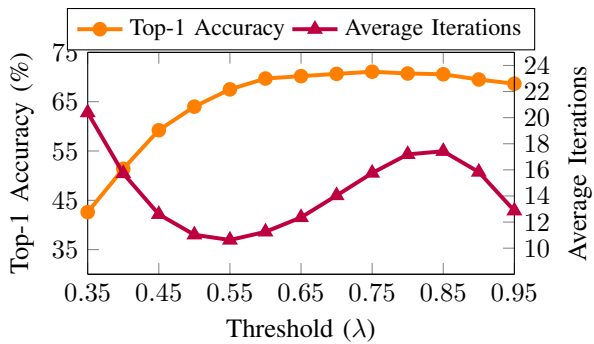
\begin{figure}[t]
\centering
\begin{tikzpicture}
  \begin{axis}[
    width=0.4\textwidth,
    height=0.25\textwidth,
    xlabel={Threshold ($\lambda$)},
    ylabel={Top-1 Accuracy (\%)},
    ymin=30, ymax=75, xmin=0.35, xmax=0.95,
    xtick={0.35, 0.45, 0.55, 0.65, 0.75, 0.85,  0.95},
    ytick={35, 45, 55, 65, 75},
    axis y line*=left,
    legend style={at={(0.5,1.2)}, anchor=north, legend columns=2, font=\small},
    legend entries={Top-1 Accuracy, Average Iterations}
  ]
    \addplot[color=orange, mark=*, line width=1.5pt] coordinates {
      (0.35,42.6)
      (0.4,51.4)
      (0.45,59.22)
      (0.5,64)
      (0.55,67.5)
      (0.6,69.7)
      (0.65,70.18)
      (0.7,70.64)
      (0.75,71.08)
      (0.8,70.72)
      (0.85,70.56)
      (0.9,69.5)
      (0.95,68.64)
    };

    \addlegendimage{mark=triangle*, color=purple, line width=1.5pt}
  \end{axis}

  \begin{axis}[
    width=0.4\textwidth,
    height=0.25\textwidth,
    xmin=0.35, xmax=0.95, ymin=8, ymax=25,
    ylabel={Average Iterations},
    ytick={10,12,14,16,18,20,22,24},
    xtick=\empty,
    axis y line*=right,
  ]
    \addplot[color=purple, mark=triangle*, line width=1.5pt] coordinates {
      (0.35,20.3852)
      (0.4,15.7218)
      (0.45,12.5938)
      (0.5,11.0262)
      (0.55,10.6266)
      (0.6,11.2508)
      (0.65,12.3548)
      (0.7,14.0356)
      (0.75,15.755)
      (0.8,17.186)
      (0.85,17.4228)
      (0.9,15.8312)
      (0.95,12.8646)
    };
  \end{axis}
\end{tikzpicture}
\caption{Effect of changing threshold ($\lambda$) on top-1 accuracy and average iterations for ResNet and Patch Size= 64.}
\label{fig:threshblack_box}
\end{figure}

\cref{fig:threshblack_box} illustrate the impact of the masking threshold, $\lambda$, on classifier accuracy and the average number of iterations until convergence. We observe that a low masking threshold leads to suboptimal accuracy, with the most effective range being around 0.75, although performance begins to plateau at approximately 0.6. The average number of iterations is notably higher at lower thresholds, as a lower threshold increases the likelihood of masking more pixels at each iteration. The lowest number of iterations occurs near a threshold of 0.55, but as the threshold increases beyond this point, the number of iterations rises, possibly because fewer pixels are masked per iteration. If the threshold is too high, the number of iterations drops, but accuracy also shows a slight decline.


\subsection{Effect of Super Resolution}
\begin{figure}[t]
    \centering
    \includegraphics[width=.9\columnwidth]{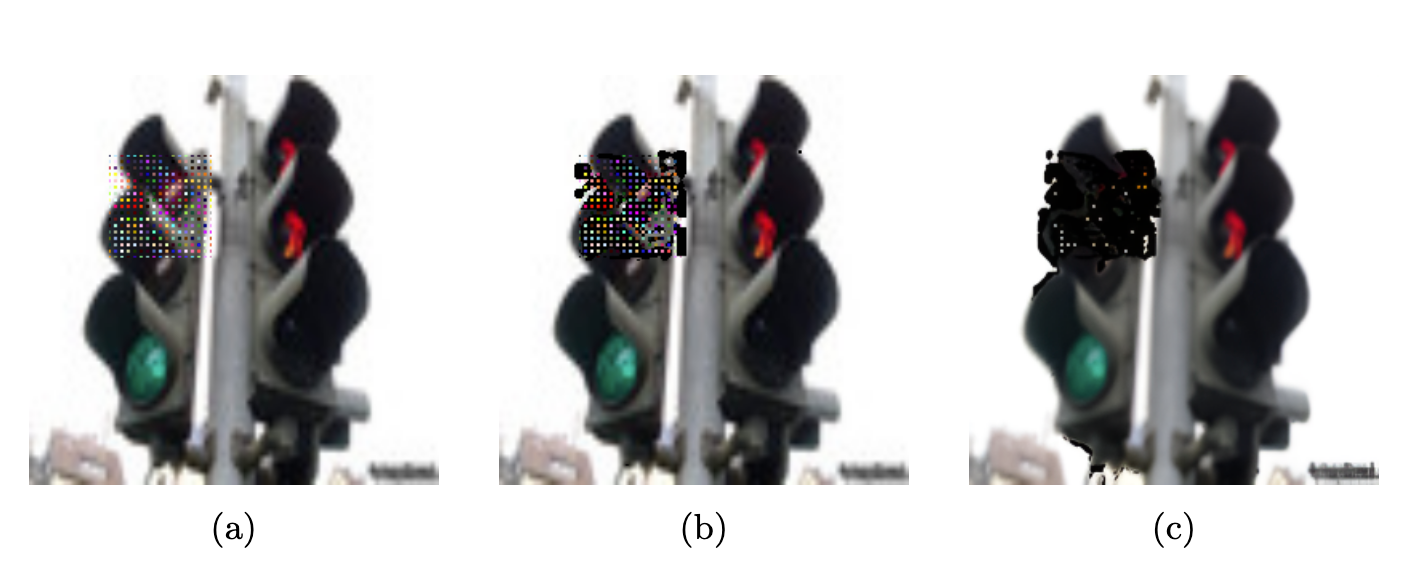}
    \caption{ (a) White-box attack with the smoothed adversarial patch. (b) Result after applying the naive defense, which fails because the adversarial patch is smoothed. (c) Output of our proposed model, successfully masking the adversarial patch.} 
    \label{fig:ours_naive}
\end{figure}

The primary reason for using a super-resolution model like Real-ESRGAN instead of a simple downsampling and upsampling operation is that GAN-based models aim to map the image distribution closer to that of natural images. In contrast, naive upsampling and downsampling merely perform basic averaging, which can be exploited by attackers. Specifically, an attacker can craft sufficiently smooth noise, so it remains unchanged after downsampling and upsampling. We demonstrate this in Figure \ref{fig:ours_naive}. In this comparison, the threshold and setup remain consistent, with the only difference being that the naive method replaces the GAN-based model with simple upsampling. The results show that, unlike our approach, the naive method fails to mask the adversarial patch effectively.

\subsection{Additional Results}

We provide additional results for different experiments, including results on the COCO (object detection) dataset, distributed white-box patches, and alternative solutions in \textbf{\textit{Appendix A}} (A1-A6).

%% file: Sections/appendix.tex
\section{Appendix}
\subsection{COCO \& DPatch Attack}
\label{sec:coco_appendix}

We evaluate \sys{} on the \textbf{COCO} dataset~\cite{lin2014microsoft} using \textbf{Faster R-CNN}~\cite{ren2015faster} for object detection, under a \textbf{DPatch} threat~\cite{chen2022dpatch}. DPatch stands for \emph{``An Adversarial Patch Attack on Object Detectors''} and we generated the patch using the framework provided by the Adversarial Robustness Toolbox~\cite{art_toolbox}. Specifically, the attack inserts a $96\times 96$ adversarial region designed to disrupt detection performance.

\subsubsection{Experimental Setup}
We do not retrain or modify the detector; instead, we apply \sys{} to each adversarial image, then feed the purified output into the original Faster R-CNN. This underscores \sys{}'s \emph{plug-and-play} capability, as no task-specific retraining is required.

\subsubsection{Results and Observations}
Table~\ref{tab:coco_dpatch} summarizes the micro-precision on randomly selected COCO pictures. Clean detection accuracy of \textbf{60\%} plunges to \textbf{35\%} under DPatch, but \sys{} restores it to \textbf{58\%}, indicating robust generalization beyond classification tasks. 

\begin{table}[h!]
  \centering
  \caption{Faster R-CNN~\cite{ren2015faster} on COCO~\cite{lin2014microsoft} with DPatch~\cite{chen2022dpatch} ($96 \times 96$).}
  \label{tab:coco_dpatch}
  \begin{tabular}{lccc}
    \toprule
    \textbf{Model} & \textbf{Clean} & \textbf{Attack} & \textbf{After \sys{}} \\
    \midrule
    Faster R-CNN & 60\% & 35\% & 58\% \\
    \bottomrule
  \end{tabular}
\end{table}

\begin{figure}[h!]
  \centering
  \includegraphics[width=0.40\textwidth]{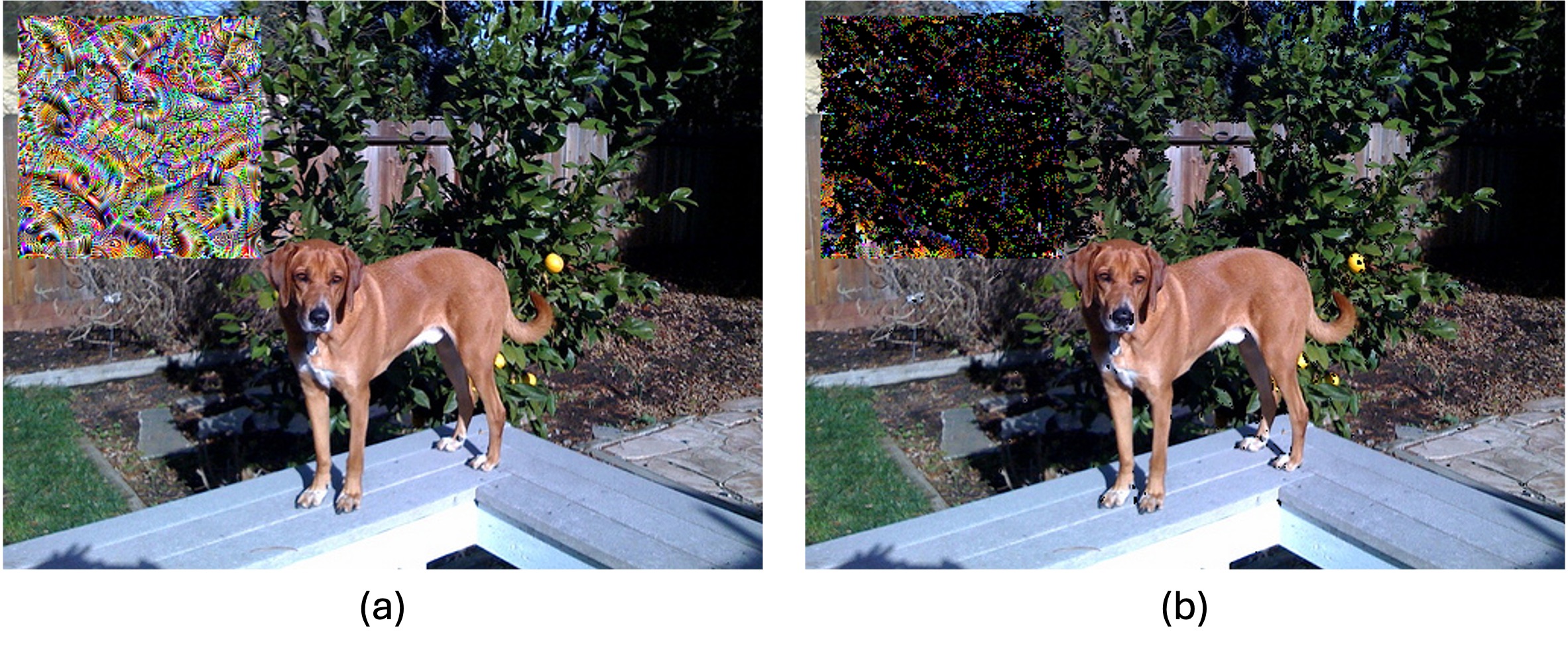} 
  \caption{Visual results on COCO: (\textbf{a}) original adversarial image (DPatch), (\textbf{b}) purified by \sys{}. The $96\times96$ patch is successfully mitigated.}
  \label{fig:coco_quality}
\end{figure}

Figure~\ref{fig:coco_quality} shows an example COCO image before and after purification. On the left (\textbf{a}), the original image with a $96 \times 96$ DPatch is shown. On the right (\textbf{b}), \sys{} effectively neutralizes the patch while minimally affecting the rest of the scene. These results confirm that \sys{} successfully purifies adversarial patches on COCO and remains applicable to diverse vision tasks.

\subsection{Distributed White-Box Patches vs. PatchCleanser}
\label{sec:distributed_patch_appendix}

In this section, we explore \emph{distributed} adversarial patches where multiple $32\times32$ regions are placed throughout the image. Crucially, the attacker has \textbf{white-box} knowledge of our \sys{} method, specifically crafting these patches to exploit \sys{}’s iterative masking. We also evaluate PatchCleanser~\cite{xiang2022patchcleanser} under the same multi-patch distribution for comparison, even though PatchCleanser itself does not operate in a white-box mode.
\subsubsection{Experimental Setup}
For each experiment, we increment the number of $32\times32$ patches scattered across the image. The total adversarial area thus becomes increasingly fragmented, posing a stronger challenge. While \sys{} faces a white-box attacker, PatchCleanser is tested as-is. This setup highlights the difference in how each defense copes with multiple small patches.

\begin{figure}
\begin{tikzpicture}
    \begin{axis}[
            width=0.45\textwidth,
    height=0.3\textwidth,
        xlabel={Number of Patches},
        ylabel={Robustness},
        xlabel style={font=\large},
        ylabel style={font=\large},
        legend style={font=\small ,at={(0.535,0.98)}, anchor=north, legend columns=-1},
        ymin=0, ymax=1,
        xtick={0,1,2,4,8},
        ytick={0,0.2,0.4,0.6,0.8,1},
        grid=both,
        grid style={dashed, gray!30},
        legend cell align={left},
        tick label style={font=\small},
        label style={font=\small}
    ]
        \addplot[color=blue, mark=*] coordinates {
            (0, 0.7986)
            (1, 0.74)
            (2, 0.6219)
            (4, 0.5049)
            (8, 0.3158)
        };
        \addlegendentry{\sysp}

        \addplot[color=teal, mark=square*] coordinates {
            (0, 0.6898)
            (1, 0.5164)
            (2, 0)
            (4, 0)
            (8, 0)
        };
        \addlegendentry{PatchCleanser~\cite{xiang2022patchcleanser}}
    \end{axis}
\label{fig:white_dorpach}
\end{tikzpicture}
\caption{Comparison of the impact of increasing the number of patches in a white-box attack between our method and Patch Cleanser. Each patch measures 32×32, with a (low) noise level of 8/255.}
\label{fig:white_dorpach}
\end{figure}
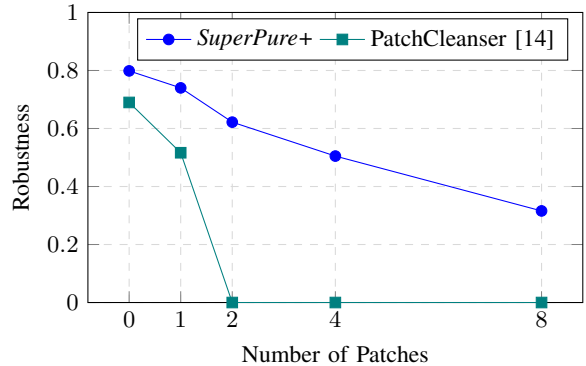

\subsubsection{Results and Observations}
Figure~\ref{fig:white_dorpach} plots the robust accuracy as the number of distributed patches increases. Despite the attacker’s full knowledge of \sys{}, our method preserves high robustness. In contrast, \textbf{PatchCleanser}~\cite{xiang2022patchcleanser} degrades rapidly as more patches are introduced.

\paragraph{Discussion.}
These experiments confirm:
\begin{itemize}
  \item \textbf{Iterative Masking Under White-Box Attacks.} Even when the attacker tailors multiple patches with inside knowledge of \sys{}, its repeated non-linear reconstructions still hinder patch survival.
  \item \textbf{PatchCleanser Limitations.} Although PatchCleanser performs well against fewer/larger patches, it fails to maintain robustness when faced with many distributed patches.
\end{itemize}
Hence, \sys{} outperforms PatchCleanser in complex, high-fragmentation adversarial scenarios. 

\subsection{Evaluation on CIFAR-10 \& CIFAR-100}
\label{sec:cifar_appendix}

We additionally tested \sys{} on the \textbf{CIFAR-10} and \textbf{CIFAR-100} datasets~\cite{krizhevsky2009learning} to assess its generalization to smaller-resolution images. We employed a ResNet-18~\cite{he2016deep} model initially pretrained on ImageNet~\cite{deng2009imagenet} and then fine-tuned for each CIFAR dataset. Since we use a $32\times32$ adversarial patch, we \textit{upscaled} each $32\times32$ CIFAR image to $256\times256$ so that the patch would not occupy the entire image, thus creating a realistic test scenario for our iterative defense.

\subsubsection{Experimental Setup and Preliminary Results}
Table~\ref{tab:cifar_results} shows the accuracy on clean CIFAR images, the accuracy under the $32\times32$ patch attack, and the recovered accuracy after applying \sys{}. Despite the aggressive upscaling, \sys{} substantially mitigates adversarial damage, suggesting that its iterative down-up masking pipeline is \emph{dataset-agnostic} and does not rely on large native resolutions.

\begin{table}[htbp]
  \centering
  \caption{CIFAR-10 and CIFAR-100 results using a fine-tuned ResNet-18 (images upscaled to $256\times256$).}
  \label{tab:cifar_results}
  \begin{tabular}{lccc}
    \toprule
    \textbf{Dataset} & \textbf{Clean} & \textbf{Attack} & \textbf{After \sys{}} \\
    \midrule
    CIFAR-10   & 94.60\% & 2.38\%  & 85.78\% \\
    CIFAR-100  & 80.09\% & 2.06\%  & 62.26\% \\
    \bottomrule
  \end{tabular}
\end{table}

These preliminary findings highlight \sys{}’s resilience against patch-based adversarial threats, even for comparatively small datasets. 

\subsection{Alternative SR: Diffusion-Based Models}
\label{sec:diffusion_appendix}

In the main paper, we selected Real-ESRGAN for super-resolution due to its relatively fast inference, which is critical for our \emph{iterative} down-up cycles. However, recent diffusion-based SR approaches, such as \textbf{SR3}~\cite{sr3ho2021image}, can sometimes yield higher-quality reconstructions. We briefly experimented with SR3 to explore this trade-off.

\subsubsection{Latency vs. Robustness Trade-off}
Our tests show that while SR3 can improve image fidelity slightly, it is \emph{significantly slower}. On a single image, SR3 may take \textit{several seconds}, making multiple passes impractical. By contrast, Real-ESRGAN performs sufficiently fast to allow repeated down-up cycles. Moreover, we observed \emph{only minor differences} in overall adversarial robustness between SR3 and Real-ESRGAN, reaffirming that \emph{any sufficiently non-linear SR} method can disrupt patch artifacts effectively, as long as it shifts images closer to the \emph{natural} manifold.

\paragraph{Conclusion.}
If speed is not a concern, a diffusion-based SR might enhance the final visual quality slightly more. However, for our repeated masking design, a lightweight and relatively fast SR generator remains more suitable for real-time or large-scale deployment.

\begin{figure}[t]
    \centering
    \includegraphics[width=.85\columnwidth]{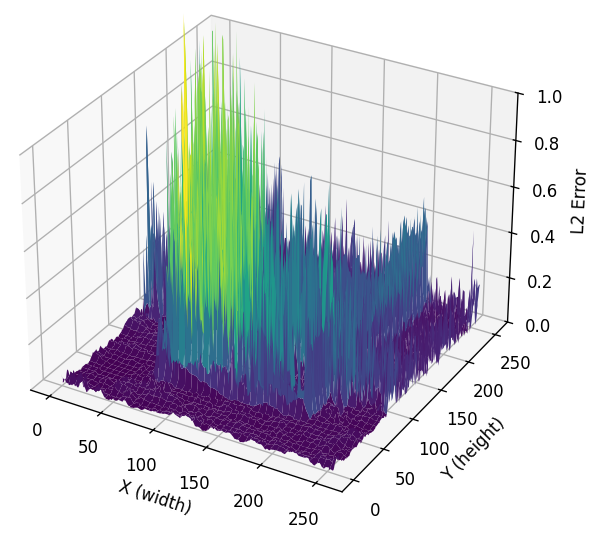}
    \caption{Reconstruction error across adversarial patch regions versus non-patch regions, illustrating that patched pixels exhibit significantly higher mean squared error (MSE) post-GAN upsampling.}
    \label{fig:patch_recon_error}
\end{figure}

\begin{table}[htbp]
\centering
\caption{Accuracy vs. Patch Size for Reversing Order of Rescaling in the Masking Phase}
\label{tab:upthendown}
{%
\begin{tabular}{lccc}
\toprule
\textbf{Patch Size} & \textbf{\sys Acc.} & \textbf{Reverse Acc.} \\
\midrule
0$\times$0   & 71.20\% & 69.66\% \\
16$\times$16 & 70.48\% & 69.90\% \\
32$\times$32 & 70.10\% & 69.52\% \\
48$\times$48 & 68.52\% & 68.30\% \\
64$\times$64 & 66.20\%  & 65.90\% \\
96$\times$96 & 58.18\%  & 58.66\% \\
\bottomrule
\end{tabular}
}
\end{table}

\subsection{Effect of Rescaling Order} 
\cref{tab:upthendown} presents the effect of reversing the rescaling order (i.e., down-up vs. up-down) during the masking phase on accuracy, evaluated at various patch sizes. Specifically, the table compares the accuracy (\sys Acc.) of \sys, which downsamples the image before super-resolution in the masking phase, against a reversed approach that upsamples first (Reverse Acc.). The results show that the performance differences are modest for all patch sizes, confirming that the similar mechanisms of the two pipelines (as outlined in \cref{sec:opt}) translate to similar performance in terms of robustness.

\subsection{Reconstruction Error Visualization}
\label{sec:appendix-recon}

In Figure~\ref{fig:patch_recon_error}, we illustrate an example image where the adversarial patch region shows distinctly higher reconstruction error than non-patch areas after GAN upsampling.